\documentclass[10pt,twocolumn,letterpaper]{article}

\usepackage{cvpr}             
\setkeys{Gin}{draft=false}
\usepackage[table]{xcolor}
\definecolor{cvprblue}{rgb}{0.21,0.49,0.74}
\usepackage[pagebackref,breaklinks,colorlinks,allcolors=cvprblue]{hyperref}

\def\paperID{*****} 
\def\confName{CVPR}
\def\confYear{2026}

\title{The First EgoCross Challenge at EgoVis 2026:\\
Cross-Domain Egocentric Video Question Answering}

\author{
Yuqian Fu$^{*}$ \and
Tianwen Qian$^{*}$ \and
Yanjun Li$^{*}$ \and
Yu Li$^{*}$ \and
Kunyu Peng$^{*}$ \and
Xu Zheng$^{*}$ \and
Yongqin Xian$^{*}$ \and
Alessio Tonioni$^{*}$ \and
Yanwei Fu$^{*}$ \and
Xiaoling Wang$^{*}$ \and
Danda Paudel$^{*}$ \and
Federico Tombari$^{*}$ \and
Luc Van Gool$^{*}$ \and 
Leyi Wu \and Yifan Zhao \and Jinjie Zhang \and Yinchuan Li \and Yingcong Chen \and Zixu Li \and Zhiwei Chen \and Zhiheng Fu \and Wenbo Wang \and Yupeng Hu \and Weili Guan \and Liqiang Nie \and Takuya Murakawa \and Toru Tamaki \and Yi Wen \and Zhenglin Du \and Zhengyang Li \and Lingling Li \and Licheng Jiao \and Wenping Ma
}

\begin{document}
\maketitle

\begingroup
\renewcommand\thefootnote{*}
\footnotetext{Yuqian Fu,
Tianwen Qian,
Yanjun Li,
Yu Li,
Kunyu Peng,
Xu Zheng,
Yongqin Xian,
Alessio Tonioni,
Yanwei Fu,
Xiaoling Wang,
Danda Paudel,
Federico Tombari, and
Luc Van Gool are the EgoCross@EgoVis26 challenge organizers. The other authors are participants in this challenge. \\ 
Appendix~\ref{sec:teams} contains all the authors’ team names and affiliations. \\
Corresponding to Yuqian Fu (yuqian.fu@kaust.edu.sa).
} 
\endgroup

\begin{abstract}
EgoCross is a cross-domain egocentric video question answering benchmark designed to evaluate whether multimodal large language models can generalize beyond common daily-life scenarios. The first EgoCross Challenge was hosted at the Third EgoVis Workshop at CVPR 2026 and evaluated models on first-person videos from four target domains: surgery, industrial assembly, extreme sports, and animal perspectives. Each test example consists of an egocentric video clip, a question, and four candidate answers, from which the model must select the correct option. This technical report introduces the challenge task, benchmark resources, and two official Codabench tracks. The Source-Limited Track restricts participants to the official baseline model and a small support set, whereas the Open-Source Track permits broader choices of models and training data under rules that prohibit the manual construction of target-domain training data. In total, the challenge received more than 1,500 submissions from over 130 participants, with 19 teams participating in the Open-Source Track and 38 teams in the Source-Limited Track. We further present the official leaderboard results and summarize the winning solutions from both tracks. We hope that this report will serve as a useful technical reference for advancing cross-domain egocentric video understanding. All resources, including the challenge data, baseline implementation, and code released by the winning teams, are made publicly available.
\end{abstract}    
\section{Introduction}
Egocentric video understanding~\cite{plizzari2024outlook,bandini2020analysis,li2026challenges,fu2024objectrelator,pan2025v,wang2025streameqa,mahdi2025exo2egosyn,zhu2026egosound,lu2026affinspace} aims to interpret the world from a first-person perspective. By capturing visual experiences from the viewpoint of the camera wearer, egocentric videos provide rich cues about human attention, intention, interactions, and task progression. These distinctive characteristics make egocentric video understanding fundamental to a wide range of applications, including wearable assistants~\cite{zhang2023accessible,janaka2024tom,wen2025snap}, embodied AI~\cite{mon2025embodied, zou2026sis, wang2025streameqa, li2025clivis}, navigation~\cite{li2026bridging}, robotics~\cite{wang2026afford, lin2026la4vla,lin2026evo,du2026focusable,li2025learning,wang2026oflow,wang2026ocra, lin2025evo0, lin2025evo1}, and augmented reality~\cite{li2026mask2iv,arena2022overview,koumpouros2024revealing}. 

\begin{figure*}[!t]
  \centering
  \vspace{-0.1in}
  \includegraphics[draft=false,width=0.99\linewidth]{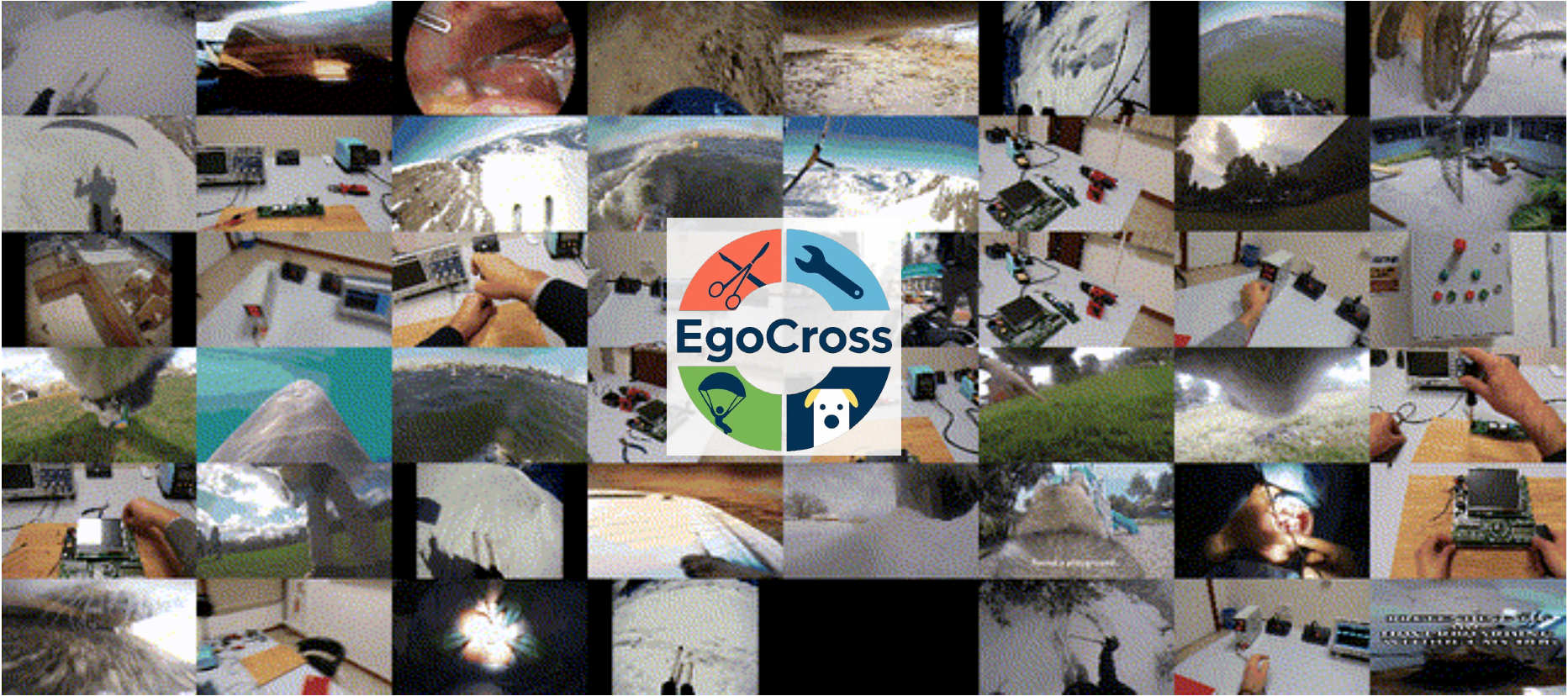}
\caption{\textbf{EgoCross task examples.} The challenge covers professional and non-daily egocentric domains, including surgery, industry, extreme sports, and animal perspectives, with each question categorized by its corresponding task type.}
  \label{fig:teaser}
    \vspace{-0.1in}
\end{figure*}

However, most existing egocentric benchmarks~\cite{damen2018scaling,fan2019egovqa,jia2022egotaskqa,mangalam2023egoschema,cheng2024egothink,plizzari2025omnia,xu2025tog,zhu2026egosound,zhang2026egonight,xu2025tog} primarily focus on common daily activities, such as cooking, household work, and routine object interactions. Although these settings provide an important foundation, real-world first-person applications often involve specialized domains with distinct visual appearance, terminology, temporal structure, and operational risks. A model that recognizes everyday objects may still fail to identify surgical instruments, understand circuit-board assembly, anticipate skiing motions, or interpret animal-mounted footage.

To investigate how models cope with these challenging yet practically important domain gaps, EgoCross~\cite{li2026egocross} extends domain adaptation research from conventional vision tasks, such as image classification~\cite{zhuo2024unified, zhuo2026segdp,fu2023styleadv,fu2021meta, zhangji2025closer, TangYLT22, zhangji2025reliable, zhang2026channel, zou2024flatten,zou2024attention,zhuo2026prompt}, object detection~\cite{li2026domain, TangLZHT25, fu2024cross,zhangji2023global,jiang2026remedying}, and semantic segmentation~\cite{wang2020classes, wang2025toward, wei2026adaptive,tong2024lightweight,liu2025devil,tong2025self}, to the more advanced setting of egocentric video question answering. Specifically, EgoCross introduces a cross-domain egocentric VQA benchmark spanning four target domains: surgery, industry, extreme sports, and animal perspectives. The benchmark contains 798 video clips and 957 question--answer pairs, providing, for the first time, a comprehensive testbed for evaluating egocentric VQA under substantial domain shifts.

As EgoCross was only recently introduced, the first \textbf{EgoCross Challenge}\footnote{EgoCross: \url{https://egocross-benchmark.github.io/}} was formally organized as part of the Third \textbf{\textit{EgoVis Workshop}} at CVPR 2026\footnote{EgoVis: \url{https://egovis.github.io/cvpr26/}} to increase its visibility within the community and stimulate further research on cross-domain egocentric VQA. EgoVis brings together representative research, benchmarks, and challenges from the egocentric vision community, making it a natural venue for assessing progress in cross-domain first-person video understanding. The challenge comprises two official Codabench competitions: the \textit{Source-Limited Track} and the \textit{Open-Source Track}. These tracks are designed to investigate complementary research questions. The Source-Limited Track emphasizes controlled and fair comparisons of adaptation methods under a fixed baseline model and a small target-domain support set. In contrast, the Open-Source Track evaluates how far stronger architectures, broader pretraining resources, and additional permissible data can advance cross-domain egocentric video question answering.

Overall, the challenge received more
than 1,500 submissions from over 130 participants, with 19
teams participating in the Open-Source Track and 38 teams
in the Source-Limited Track.  This technical report provides a comprehensive overview of the first EgoCross Challenge. Section~\ref{sec:challenge} describes the challenge design, including the task formulation, competition tracks, dataset composition, baseline model, and evaluation protocol. Section~\ref{sec:results} reports and analyzes the official leaderboard results. Section~\ref{sec:methods} summarizes the winning solutions proposed by participating teams in both the Open-Source and Source-Limited Tracks.

\section{EgoCross Challenge}
\label{sec:challenge}

\subsection{Challenge Overview}
\label{sec:overview}
Building on the EgoCross benchmark introduced above, the challenge aims to evaluate and promote cross-domain egocentric video understanding. We adopt a closed-ended video question answering setting, referred to as \emph{CloseQA}, in which each example consists of an egocentric video clip, a natural-language question, and four candidate answers. Participants are required to select the correct answer for each example.
Note that the original EgoCross benchmark~\cite{li2026egocross} provides both closed-ended and open-ended question answering formats for each question. However, considering the computational cost of evaluation and the reliability of automatic metrics, the challenge focuses exclusively on the CloseQA setting.

\begin{figure}[!t]
  \centering
  \vspace{-0.1in}
  \includegraphics[draft=false,width=0.99\linewidth]{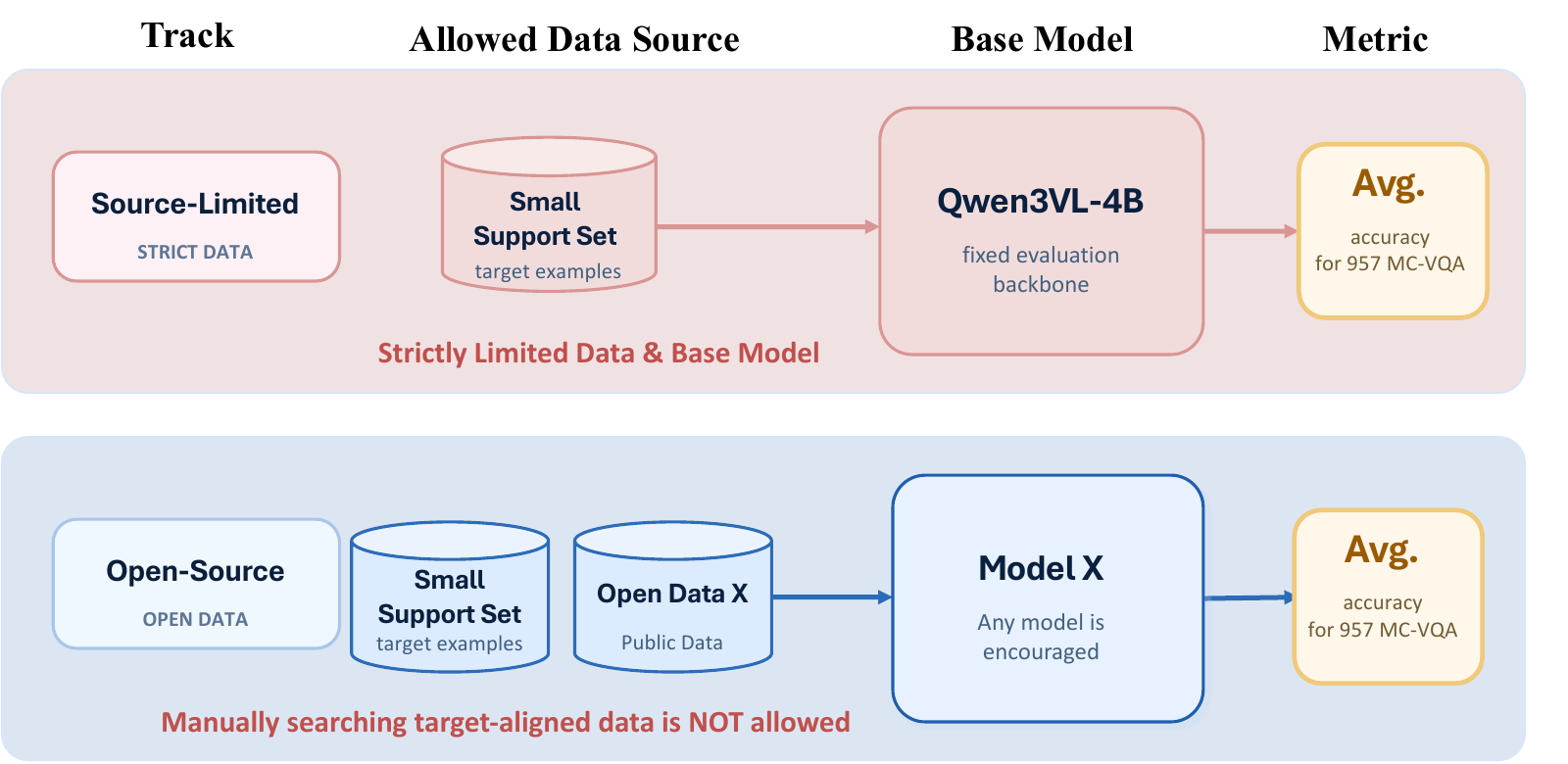}
\caption{\textbf{Overview of the two EgoCross challenge tracks.}
The Source-Limited Track restricts participants to the official Qwen3VL-4B backbone and a small target-domain support set, enabling controlled comparison under limited data and model resources. The Open-Source Track allows participants to use alternative models and additional public data, while prohibiting the manual collection of target-aligned training data. Both tracks are evaluated by average accuracy over 957 multiple-choice VQAs.}
  \label{fig:setup}
    \vspace{-0.1in}
\end{figure}

The challenge includes two official Codabench tracks: the \emph{Source-Limited Track}\footnote{Source-Limited Track: \url{https://www.codabench.org/competitions/11279/}}, which restricts participants to the official baseline model and provides a small, fixed support set for model adaptation; and the \emph{Open-Source Track}\footnote{Open-Source Track: \url{https://www.codabench.org/competitions/13868/}}, which allows broader choices of models and training data while prohibiting the manual construction of target-domain training data. Fig.~\ref{fig:setup} illustrates the key differences between the two tracks.  

\subsection{Task Formulation}
\label{sec:task}

Given an egocentric video clip \(V\), a question \(q\), and four candidate answers
\(\mathcal{A}=\{a_A,a_B,a_C,a_D\}\), a model predicts an option
\(\hat{y}\in\{A,B,C,D\}\). The prediction is correct if \(\hat{y}=y\), where
\(y\) denotes the ground-truth option.

Although the four-choice setting has a random baseline of \(25\%\), the task remains challenging due to domain-specific distractors and the need for temporal, spatial, procedural, and fine-grained visual reasoning. The averaged accuracy is used as the metric: 
\begin{equation}
  \mathrm{Acc} = \frac{1}{N}\sum_{i=1}^{N}\mathbb{1}[\hat{y}_i=y_i],
\end{equation}
where \(N\) is the number of test questions. Predictions are submitted to Codabench in JSON format, and both overall and per-domain accuracies are reported for surgery, industry, extreme sports, and animal perspective.

\subsection{Resources: Data and Baseline}
\label{sec:resources}

\noindent\textbf{Data.}
In addition to the original EgoCross test set~\cite{li2026egocross}, we construct a small support set to facilitate model adaptation, particularly for the Source-Limited Track. The support set contains 80 multiple-choice QA samples, with 20 samples from each of the four target domains, covering a total of 1{,}259 video frames. This limited in-domain supervision encourages the development of lightweight and data-efficient adaptation methods.

\noindent\textbf{Baseline.}
We provide SFT-Qwen3VL as the official baseline for the Source-Limited Track and as a reference method for the Open-Source Track. The baseline is built upon Qwen3-VL-4B-Instruct~\cite{bai2025qwen3} and fine-tuned on the 80-sample support set using LLaMA-Factory~\cite{zheng2024llamafactory}. Full-parameter supervised fine-tuning is conducted on \(4\times\) H200 GPUs with a per-device batch size of 1 and gradient accumulation over 8 steps, resulting in an effective batch size of 32. The model is trained for 2 epochs with a learning rate of \(1\times10^{-5}\), a cosine learning-rate scheduler, and a warmup ratio of 0.1.

\noindent\textbf{Baseline Performance.}
Table~\ref{tab:baseline} reports the baseline accuracy across the four target domains. SFT-Qwen3VL achieves an average accuracy of 46.08\%, compared with 45.14\% for the zero-shot Qwen3-VL-4B model. The modest improvement highlights the difficulty of adapting to cross-domain egocentric videos using only 80 support samples and motivates the development of more effective adaptation strategies.

All challenge resources, including the support set and baseline implementation, are publicly available.\footnote{Resources: \url{https://github.com/LiYu0524/EgoCross_SFT_qwen3vl4b}}

\begin{table}[t]
  \centering
  \small
  \setlength{\tabcolsep}{4pt}
  \caption{\textbf{Baseline performance on EgoCross.} Accuracy (\%) across four target domains.}
  \label{tab:baseline}
  \begin{tabular}{@{}lccccc@{}}
    \toprule
    \textbf{Method} & \textbf{Surg.} & \textbf{Indus.} & \textbf{XSports} & \textbf{Animal} & \textbf{Avg.} \\
    \midrule
    ZSL Qwen3VL & 48.41 & 34.29 & 44.72 & 55.19 & 45.14 \\
    SFT-Qwen3VL & 47.70 & 35.10 & 48.37 & 55.19 & \textbf{46.08} \\
    \bottomrule
  \end{tabular}
\end{table}

\section{Challenge Results}
\label{sec:results}

\begin{table*}[t]
  \centering
  \small
  \setlength{\tabcolsep}{4pt}
  \caption{\textbf{Top-10 leaderboard results of the EgoCross Challenge.}
  Accuracy (\%) across four target domains and overall performance.
  The top three teams in each track are highlighted in bold.}
  \label{tab:challenge_results}
  \small
  \setlength{\tabcolsep}{8pt}
  \renewcommand{\arraystretch}{1.08}

  \resizebox{\textwidth}{!}{%
  \begin{tabular}{clccccc}
    \toprule
    \multicolumn{7}{c}{\textbf{Source-Limited Track}} \\
    \midrule
    \textbf{Rank} & \textbf{Team} & \textbf{Surgery} &
    \textbf{Industry} & \textbf{XSports} &
    \textbf{Animal Perspective} & \textbf{Overall Accuracy} \\
    \midrule
    \textbf{1} & \textbf{DomainWiseInfer} &
    \textbf{65.72} & 64.49 & \textbf{63.41} & \textbf{77.05} & \textbf{66.98} \\

    \textbf{2} & \textbf{OmniEgo-R2} &
    58.66 & \textbf{81.22} & 54.47 & 74.32 & \textbf{66.35} \\

    \textbf{3} & \textbf{TokenInj-RAA} &
    59.01 & 61.22 & 51.22 & 74.86 & \textbf{60.61} \\

    \midrule
    4  & zhengsr     & 53.00 & 61.63 & 51.63 & \textbf{77.05} & 59.46 \\
    5  & dghf2022    & 61.84 & 63.67 & 47.97 & 62.84 & 58.93 \\
    6  & zr-zhou     & 53.36 & 61.63 & 49.59 & 72.13 & 58.10 \\
    7  & Ego101      & 52.30 & 56.73 & 55.69 & 71.58 & 57.99 \\
    8  & xiang\_li   & 54.06 & 55.10 & 55.28 & 71.58 & 57.99 \\
    9  & lalala111   & 56.54 & 52.65 & 48.78 & 76.50 & 57.37 \\
    10 & odsigjoasd  & 47.00 & 60.41 & 49.59 & 73.22 & 56.11 \\

    \midrule
    \multicolumn{7}{c}{\textbf{Open-Source Track}} \\
    \midrule
    \textbf{Rank} & \textbf{Team} & \textbf{Surgery} &
    \textbf{Industry} & \textbf{XSports} &
    \textbf{Animal Perspective} & \textbf{Overall Accuracy} \\
    \midrule
    \textbf{1} & \textbf{DomainWiseInfer} &
    65.72 & 64.49 & \textbf{63.41} & 77.05 & \textbf{66.98} \\

    \textbf{2} & \textbf{OmniEgo-R2} &
    58.66 & \textbf{82.86} & 54.47 & 74.32 & \textbf{66.77} \\

    \textbf{3} & \textbf{Reflective Dialogue} &
    \textbf{74.91} & 59.18 & 52.44 & \textbf{79.23} & \textbf{65.94} \\

    \midrule
    4  & anyone           & 69.26 & 61.63 & 59.35 & 74.32 & 65.73 \\
    5  & cola\_lover      & 70.32 & 52.65 & 50.41 & \textbf{79.23} & 62.38 \\
    6  & zhengsr          & 60.07 & 61.63 & 51.63 & 77.05 & 61.55 \\
    7  & handsomeboy\_hun & 71.38 & 50.20 & 46.75 & 77.05 & 60.71 \\
    8  & TokenInj-RAA     & 59.01 & 61.22 & 51.22 & 74.86 & 60.61 \\
    9  & dghf2022         & 61.84 & 63.67 & 47.97 & 62.84 & 58.93 \\
    10 & lalala111        & 55.83 & 49.80 & 46.75 & 76.50 & 55.90 \\
    \bottomrule
  \end{tabular}%
  }
\end{table*}

The first EgoCross Challenge attracted more than 130 registered participants and received over 1{,}500 submissions across the two competition tracks. Among them, 38 teams participated in the Source-Limited Track, while 19 teams participated in the Open-Source Track. Table~\ref{tab:challenge_results} presents the top-10 teams from each track.

In the Source-Limited Track, \emph{DomainWiseInfer} ranked first with an overall accuracy of 66.98\%, followed by \emph{OmniEgo-R2} with 66.35\% and \emph{TokenInj-RAA} with 60.61\%. The winning result improves upon the official SFT-Qwen3VL baseline by 20.90 percentage points, demonstrating that substantial gains can be achieved even under tightly constrained model and data settings. The Open-Source Track exhibits a more competitive leading group. \emph{DomainWiseInfer} again achieved the highest overall accuracy of 66.98\%, closely followed by \emph{OmniEgo-R2} at 66.77\% and \emph{Reflective Dialogue} at 65.94\%. The gap between the first- and third-ranked teams is only 1.04 percentage points. 

Across both tracks, animal-perspective questions generally receive higher scores, whereas surgery and extreme-sports questions remain comparatively challenging for most teams. The substantial variation across domains further indicates that strong average performance does not necessarily imply uniform cross-domain generalization. The approaches developed by the award-winning teams, including their model selection, adaptation strategies, and inference techniques, are
described in detail in Section~\ref{sec:methods}.

\begin{figure}[h!]
    \centering
    \includegraphics[width=1.\linewidth]{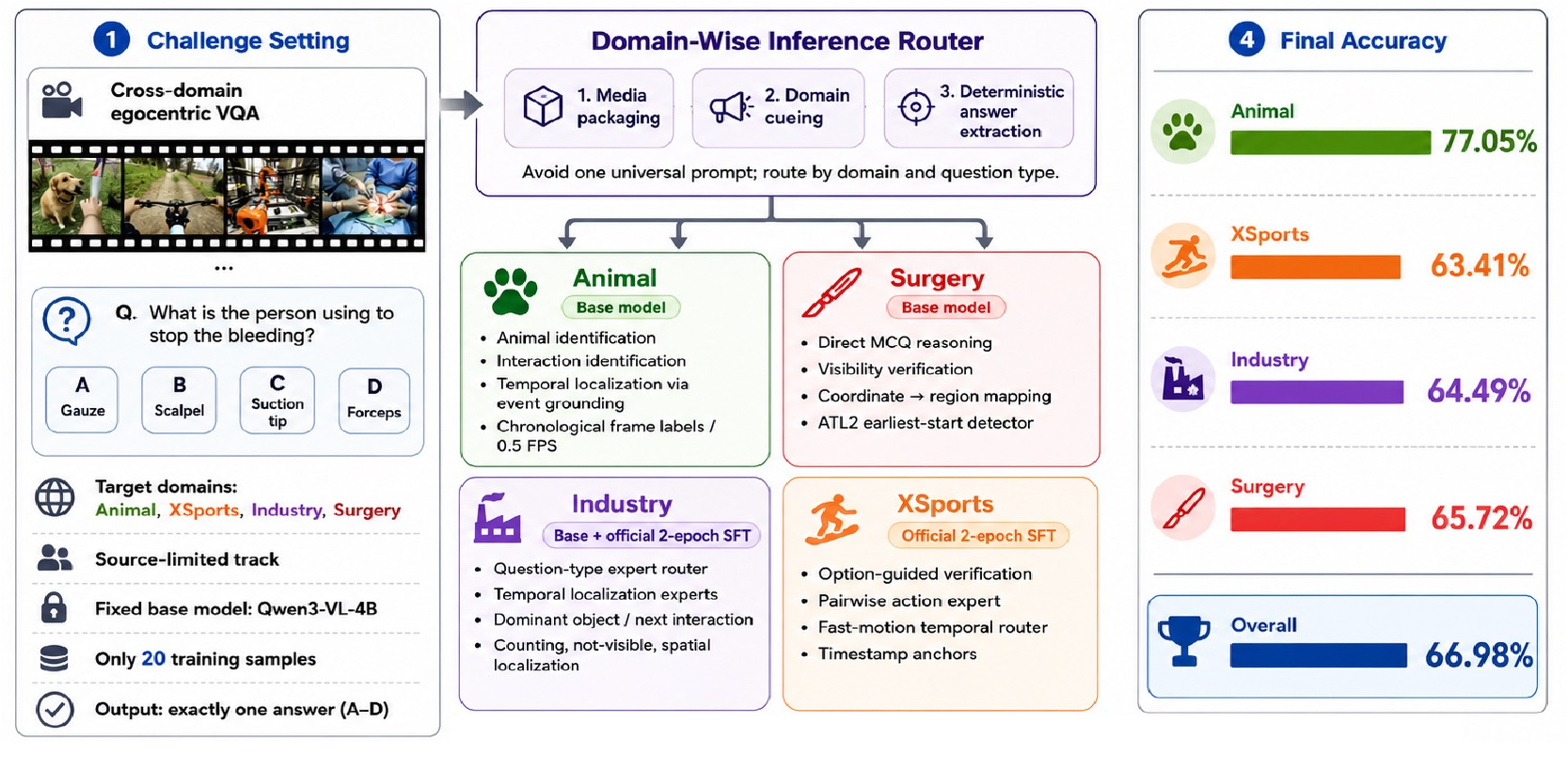}
    \caption{\textbf{Overview of DomainWiseInfer.}
    The method uses a nearly training-free, domain-wise inference framework with tailored input packaging, domain cueing, and answer extraction. }
    \label{fig:domainwiseinfer}
\end{figure}

\section{Methods}
\label{sec:methods}

This section summarizes the solutions from the award-winning teams. Rather than reproducing the full technical details of the original reports, we focus on the key designs most relevant to the EgoCross Challenge, including temporal grounding, domain-aware adaptation, structured reasoning, contextual knowledge utilization, and robust inference.

\subsection{DomainWiseInfer}
\label{sec:domainwiseinfer}

\begin{figure*}[t]
    \centering
    \includegraphics[width=\linewidth]{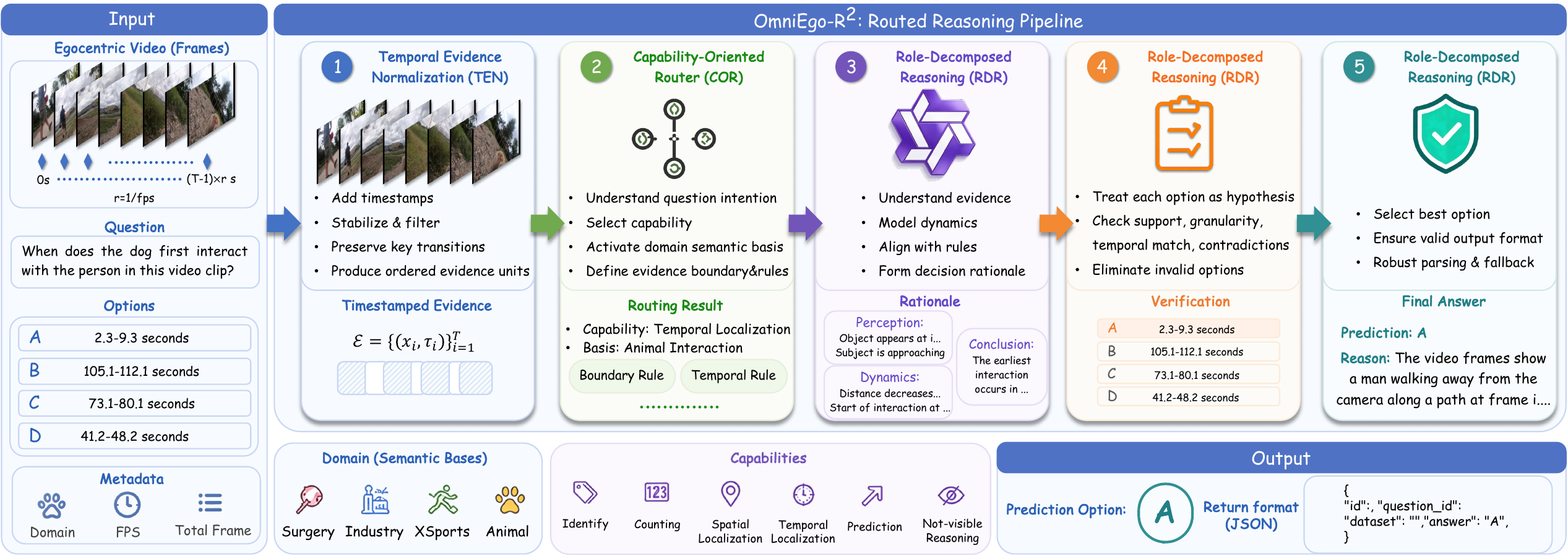}
    \caption{\textbf{Overview of OmniEgo-R$^2$~\cite{OmniEgo-R}.}
    The framework transforms each egocentric VQA sample into a unified routed
    reasoning pipeline consisting of Temporal Evidence Normalization,
    Capability-Oriented Routing, Role-Decomposed Reasoning, and Boundary-Aware
    Option Verification. Domain-specific
    semantics are incorporated as interchangeable semantic bases within the
    shared reasoning architecture.}
    \label{fig:omniego_r2}
\end{figure*}

DomainWiseInfer\footnote{Code: \url{https://github.com/YUEVII/Egocross-Challenge}}~\cite{wu2026right}, which ranked first in both the Source-Limited and Open-Source Tracks, proposes a nearly training-free domain-wise inference framework for cross-domain egocentric VQA. Rather than applying a single prompting and inference pipeline to all questions, the method routes samples according to their target domain and question characteristics, and then applies domain-specific strategies for input packaging, task cueing, and answer extraction. The approach keeps the model backbone largely unchanged: the base Qwen3-VL-4B model is directly used for the animal and surgery domains, while the official two-epoch SFT checkpoint is used for the industry and XSports domains.

As illustrated in Fig.~\ref{fig:domainwiseinfer}, the central idea is to avoid a universal inference interface and instead expose the most relevant visual and temporal cues for each domain. The framework is built around three key principles: \emph{media packaging}, which organizes visual evidence into a form more suitable for the model; \emph{domain cueing}, which highlights the most relevant domain-specific information; and \emph{deterministic answer extraction}, which ensures stable prediction in the required multiple-choice format.

For animal-perspective videos, the method emphasizes interaction grounding and chronological frame labeling to reduce ambiguities caused by unstable camera motion and salient background objects. For surgery, it uses specialized procedures for direct multiple-choice reasoning, visibility verification, spatial grounding, and temporal localization. For industry, it adopts a question-type expert router that separately handles temporal localization, object interaction, counting, not-visible identification, and spatial localization. For XSports, it introduces action-aware verification and fast-motion temporal reasoning modules. By decomposing difficult multiple-choice questions into more explicit verification, grounding, and comparison subproblems, the method improves the effectiveness of the fixed backbone under severe domain shift.

Without introducing a larger model or extensive additional training, DomainWiseInfer achieves 66.98\% overall accuracy, demonstrating that carefully designed inference interfaces can substantially improve cross-domain egocentric video question answering under constrained source-limited settings.

\subsection{OmniEgo-R$^2$}
\label{sec:omniego_r2}
OmniEgo-R$^2$\footnote{Code: \url{https://github.com/Lee-zixu/OmniEgo-R2}}~\cite{OmniEgo-R}, which ranked second in both the Source-Limited and
Open-Source Tracks, proposes a unified routed reasoning framework for
cross-domain egocentric VQA. Rather than treating EgoCross as a conventional
end-to-end multiple-choice task, the method decomposes prediction into temporal
evidence normalization, capability routing, structured reasoning, option
verification, and answer calibration. It builds upon the official
domain-specific Qwen3-VL-4B-SFT checkpoints and augments them with lightweight
test-time reasoning and robust output-parsing procedures.

\begin{figure*}[t!]
    \centering
    \includegraphics[width=\linewidth]{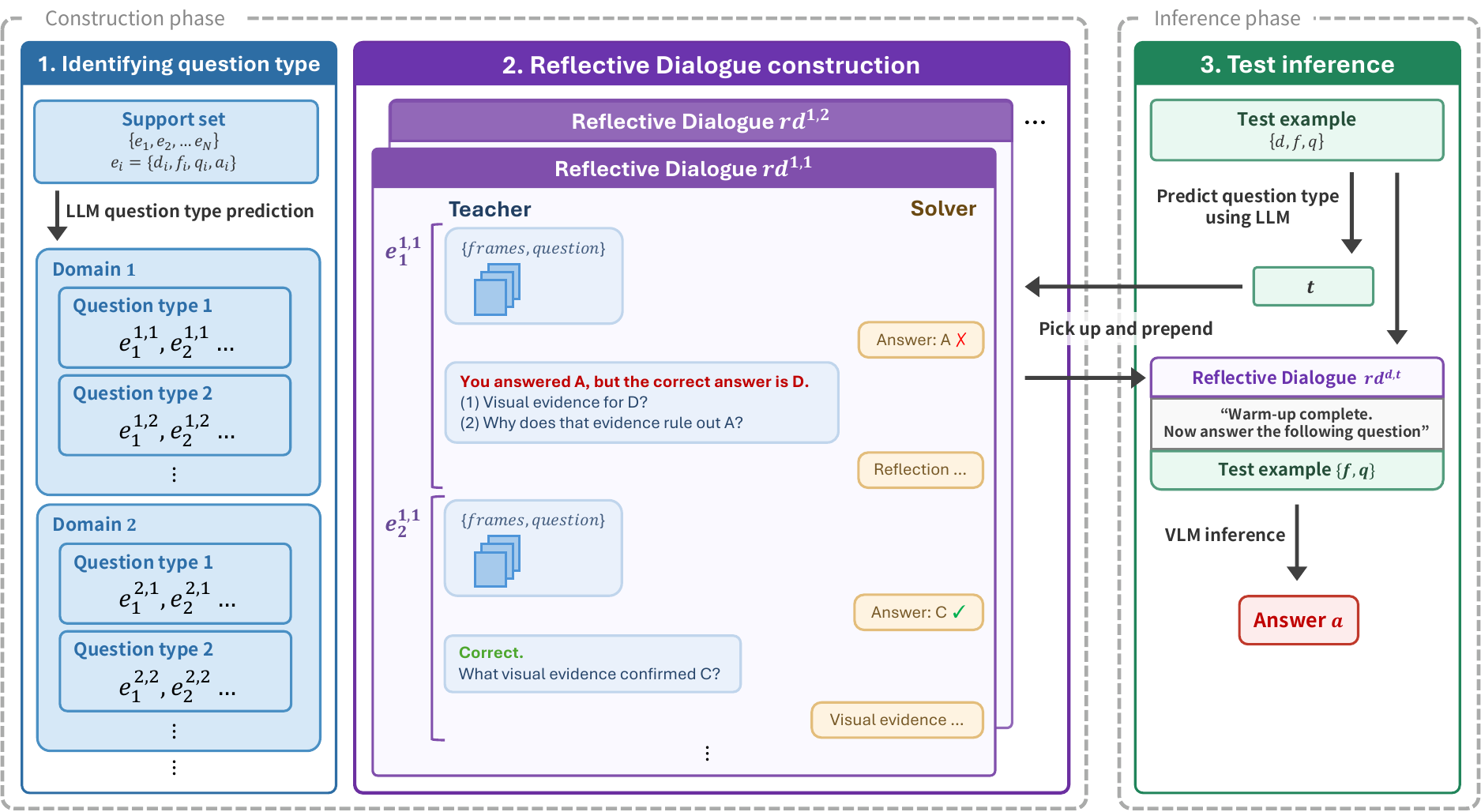}
    \caption{\textbf{Overview of Reflective Dialogue~\cite{murakawa2026reflective}.}
  Support examples are grouped by domain and question type and converted into multi-turn conversations between a Teacher and a Solver. These dialogues contain correctness feedback, visual grounding explanations, and contrastive reflections on incorrect predictions. At inference time, the corresponding dialogue is retrieved and prepended to the test question as static context.}
    \label{fig:reflective_dialogue}
\end{figure*}

As illustrated in Fig.~\ref{fig:omniego_r2}, the pipeline begins with
\emph{Temporal Evidence Normalization} (TEN), which converts sampled frames
into timestamped and chronologically ordered evidence units and guides the
model to prioritize stable, question-relevant observations. The
\emph{Capability-Oriented Router} (COR) then identifies the capability required
by the question, including identification, counting, spatial localization,
temporal localization, prediction, and not-visible reasoning. COR combines
this shared capability space with distinct domain semantic bases: tool-centric
reasoning for surgery, object-centric procedural reasoning for industry,
physics-centric embodied reasoning for XSports, and self--other behavioral
reasoning for animal-perspective videos.

The routed evidence is subsequently processed by
\emph{Role-Decomposed Reasoning} (RDR), which separates perception, temporal
or behavioral dynamics, and option-oriented decision-making to reduce premature
commitment to plausible answers. The reasoning depth is adapted to the task,
with compact expert verification used for fine-grained or controlled-vocabulary
cases. Each candidate is then treated as a hypothesis by
\emph{Boundary-Aware Option Verification} (BOV), which evaluates visual
support, semantic granularity, temporal compatibility, and potential
contradictions. Finally, \emph{Defensive Answer Calibration} (DAC) maps the
verified decision into a valid option label through robust parsing and fallback
rules.

OmniEgo-R$^2$ achieves 66.35\% overall accuracy in the Source-Limited Track
and 66.77\% in the Open-Source Track. Its strongest performance is observed in
the industry domain, reaching 81.22\% and 82.86\% in the two tracks,
respectively. These results show that a shared reasoning program can effectively
handle heterogeneous domains and capabilities when combined with explicit
temporal evidence, domain-specific semantic bases, option-level verification,
and defensive output calibration.

\subsection{Reflective Dialogue}
\label{sec:reflective_dialogue}

Reflective Dialogue\footnote{Code: \url{https://github.com/tamaki-lab/EgoCross-Reflective-Dialogue}}~\cite{murakawa2026reflective}, the third-place solution in the Open-Source Track,
proposes a training-free inference-time adaptation strategy for cross-domain egocentric VQA. Instead of directly presenting support examples as conventional in-context demonstrations, the method transforms them into structured multi-turn conversations between two agents: a \emph{Teacher}, which presents questions and provides correctness feedback, and a \emph{Solver}, which predicts answers and verbalizes the corresponding visual evidence. For incorrect predictions, the Solver additionally performs contrastive reflection by explaining both why the correct option is supported and why its original answer was invalid. The resulting dialogues are constructed offline and reused
as static context during test-time inference.

As illustrated in Fig.~\ref{fig:reflective_dialogue}, the method comprises a construction phase and an inference phase. During construction, each support question is first associated with its question type, using the provided annotation when available or an LLM-based predictor otherwise. The support examples are then grouped by domain and question type. For every example, the Teacher presents the video frames and question, the Solver predicts an answer, and the Teacher provides correctness feedback. Correct predictions trigger a
visual grounding explanation, whereas incorrect predictions trigger both
correct-answer grounding and an analysis of the original error. The resulting four-turn exchanges are sequentially concatenated into a continuous reflective dialogue for each domain--question-type pair.

At inference time, the question type of a test example is identified, and the reflective dialogue corresponding to the same domain and question type is retrieved and prepended to the test question. A separator instruction marks the end of the dialogue before the model answers the new example. Unlike iterative self-reflection methods, Reflective Dialogue does not repeatedly solve the same test question. Instead, it transfers reflective knowledge across support examples and reuses the constructed dialogue across multiple test samples. This enriches standard in-context learning with correctness feedback,
contrastive error analysis, and verbalized visual evidence without requiring task-specific parameter updates.

Using Gemini~\cite{team2023gemini} 3.1 Pro Preview with reflective dialogues and timestamped frames, the method achieves 65.94\% overall accuracy and ranks third in the Open-Source Track. The results indicate that reflective context provides benefits beyond directly supplying support-set question--answer pairs, particularly for domains involving complex procedural or fine-grained visual reasoning, while the gains are less consistent for animal-perspective videos dominated by direct visual appearance cues.

\subsection{TokenInj-RAA}
\label{sec:tokeninj_raa}

\begin{figure}[t]
    \centering
    \includegraphics[width=\linewidth]{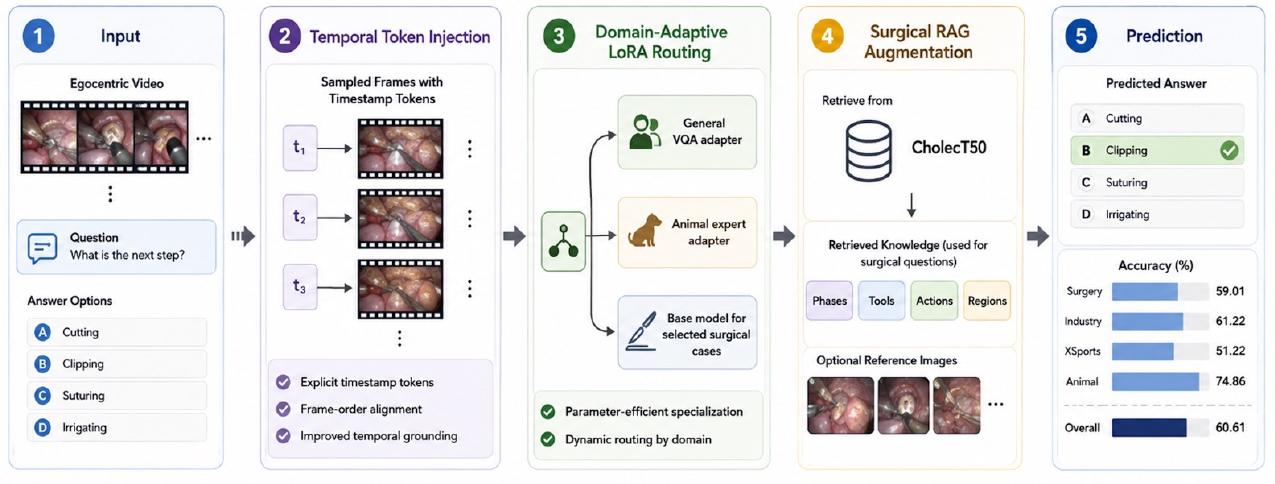}
\caption{\textbf{Overview of TokenInj-RAA.}
The method integrates timestamp-aware frame encoding, domain-adaptive LoRA routing, and CholecT50-based retrieval augmentation to improve temporal reasoning and specialized surgical understanding in cross-domain egocentric VQA.}
    \label{fig:tokeninj}
\end{figure}

TokenInj-RAA\footnote{Code: \url{https://github.com/guyi520421-byte/egoCrossCompetiton}}, which ranked third in the Source-Limited Track, proposes a modular cross-domain egocentric VQA framework combining timestamp-aware visual
modeling, parameter-efficient domain adaptation, and retrieval-augmented
reasoning, as in Fig.~\ref{fig:tokeninj}. The method is built upon Qwen3-VL-4B-Instruct and uses LoRA adapters to specialize the model under limited supervision. Rather than relying on a single uniform inference configuration, it dynamically selects the appropriate
adapter or base-model mode according to the input domain and supplements
specialized surgical questions with externally retrieved domain knowledge.

A central component of the method is explicit temporal token injection. Each sampled frame is preceded by a timestamp anchor derived from its original frame index and dataset-specific sampling rate. These temporal markers provide the model with an explicit representation of event order and elapsed time, which is particularly important for temporal localization and action-transition questions. To adapt the model efficiently, the base Qwen3-VL-4B-Instruct parameters are frozen and quantized, while LoRA modules are inserted into the attention and feed-forward projections. The model is trained to predict only the multiple-choice answer token, enabling lightweight adaptation to EgoCross-style VQA.

The method further adopts domain-adaptive inference through adapter routing. A general VQA adapter and an animal-specific expert adapter are selected according to the source domain, while selected surgical datasets can revert to the original base model when preserving its zero-shot visual representations is more effective than applying a specialized adapter. For surgical questions, TokenInj-RAA additionally introduces a lexical and multimodal retrieval pipeline based on annotated CholecT50 examples. The retrieved context includes
surgical phases, instrument categories, action triplets, spatial regions, and optionally reference images. These examples are explicitly marked as external reference knowledge to prevent them from being confused with the current test video.

By combining temporal grounding, parameter-efficient specialization, and
domain-specific knowledge retrieval, TokenInj-RAA achieves 60.61\% overall accuracy in the Source-Limited Track. Its results suggest that explicit frame-time alignment and lightweight domain adaptation can provide substantial gains, while retrieval from structured surgical annotations offers additional support for specialized medical reasoning.

\section{Conclusion}
\label{sec:conclusion}

This report presented the first EgoCross Challenge at the Third EgoVis Workshop, CVPR 2026, evaluating cross-domain egocentric video question answering across surgery, industrial assembly, extreme sports, and animal perspectives under source-limited and open-source settings. The challenge results highlight the value of domain-aware inference, structured reasoning, reflective in-context adaptation, temporal grounding, parameter-efficient fine-tuning, and retrieval augmentation, while also showing that robust generalization across specialized egocentric domains remains challenging. We hope that EgoCross and the released challenge resources will support continued progress in cross-domain first-person video understanding.

\newpage

\appendix
\section{Teams and affiliations}
\label{sec:teams}
\subsection*{EgoCross 2026 Organizer team}
\noindent\textit{\textbf{Title: }} The First EgoCross Challenge at EgoVis 2026: Cross-Domain Egocentric Video Question Answering\\
\noindent\textit{\textbf{Members: }} \\
Yuqian Fu$^{1}$ (\href{mailto:yuqian.fu@kaust.edu.sa}{yuqian.fu@kaust.edu.sa}),\\
Tianwen Qian$^{2}$ (\href{mailto:twqian@cs.ecnu.edu.cn}{twqian@cs.ecnu.edu.cn}),\\
Yanjun Li$^{2}$ (\href{mailto:51265901098@stu.ecnu.edu.cn}{51265901098@stu.ecnu.edu.cn}),\\
Yu Li$^{3}$ (\href{mailto:liyu24@m.fudan.edu.cn}{liyu24@m.fudan.edu.cn}),\\
Kunyu Peng$^{4}$,\\
Xu Zheng$^{5}$,\\
Yongqin Xian$^{6}$,\\
Alessio Tonioni$^{6}$,\\
Yanwei Fu$^{3}$,\\
Xiaoling Wang$^{2}$, \\
Danda Paudel$^{7}$,\\
Federico Tombari$^{6}$,\\
Luc Van Gool$^{7}$

\noindent\textit{\textbf{Affiliations: }}\\
$^1$ KAUST\\
$^2$ East China Normal University \\
$^3$ Fudan University \\
$^4$ KIT \\
$^5$ Great Bay University \\
$^6$ Google \\
$^7$ INSAIT\\

\subsection*{DomainWiseInfer}
\noindent\textit{\textbf{Title:}} The Right Inference Strategy Is All You Need: Nearly Training-Free Domain-Wise Inference for EgoCross Challenge\\
\noindent\textit{\textbf{Result: }} 
1st place, Source-Limited and Open-Source tracks\\
\noindent\textit{\textbf{Members: }} \\
Leyi Wu\textsuperscript{\rm 1,3,$*$} (\href{mailto:lwu398@connect.hkust-gz.edu.cn}{lwu398@connect.hkust-gz.edu.cn}), \\
Yifan Zhao\textsuperscript{\rm 1,$*$}(\href{mailto:yzhao642@connect.hkust-gz.edu.cn}{yzhao642@connect.hkust-gz.edu.cn}), \\
Jinjie Zhang\textsuperscript{\rm 1, $*$}(\href{mailto:jzhang103@connect.hkust-gz.edu.cn}{jzhang103@connect.hkust-gz.edu.cn}), \\
Yinchuan Li\textsuperscript{\rm 3}, \\
Yingcong Chen\textsuperscript{\rm 1,2,$\dag$}(\href{mailto:yingcongchen@ust.hk}{yingcongchen@ust.hk}), \\
\noindent\textit{\textbf{Affiliations: }}\\
$^1$ HKUST(GZ)\\
$^2$ HKUST \\
$^3$ Knowin \\

\subsection*{OmniEgo-R$^2$}
\noindent\textit{\textbf{Title:}} OmniEgo-R$^2$: A Routed Reasoning Framework for the 1st Cross-Domain EgoCross Challenge at CVPR 2026\\
\noindent\textit{\textbf{Result:}} 2nd place, Source-Limited and Open-Source tracks\\
\noindent\textit{\textbf{Members: }} \\
Zixu Li$^{1}$ (\href{mailto:lizixu.cs@gmail.com}{lizixu.cs@gmail.com}), \\
Zhiwei Chen$^1$(\href{mailto:zivczw@gmail.com}{zivczw@gmail.com}), \\
Zhiheng Fu$^1$(\href{mailto:fuzhiheng8@gmail.com}{fuzhiheng8@gmail.com}), \\
Wenbo Wang$^{1}$ (\href{mailto:wangwenbo@mail.sdu.edu.cn}{wangwenbo@mail.sdu.edu.cn}), \\
Yupeng Hu$^{1}$(\href{mailto:huyupeng@sdu.edu.cn}{huyupeng@sdu.edu.cn}), \\
Weili Guan$^{2}$  (\href{mailto:honeyguan@gmail.com}{honeyguan@gmail.com}), \\
Liqiang Nie$^2$ (\href{mailto:nieliqiang@gmail.com}{nieliqiang@gmail.com}) \\
\noindent\textit{\textbf{Affiliations: }} \\
$^1$ Shandong University \\
$^2$ Harbin Institute of Technology (Shenzhen)\\

\subsection*{Reflective Dialogue}
\noindent\textit{\textbf{Title: }} Reflective Dialogue between Teacher and Solver Agents for Video Question Answering \\
\noindent\textit{\textbf{Result: }} 3rd place, Open-Source track\\
\noindent\textit{\textbf{Members: }} \\
Takuya Murakawa$^{1}$ (\href{mailto:t.murakawa.080@nitech.jp}{t.murakawa.080@nitech.jp})  \\
Toru Tamaki$^{1}$ (\href{mailto:tamaki.toru@nitech.ac.jp}{tamaki.toru@nitech.ac.jp})\\
\noindent\textit{\textbf{Affiliations: }} \\
$^{1}$ Nagoya Institute of Technology\\

\subsection*{TokenInj-RAA}
\noindent\textit{\textbf{Title: }} Enhancing Visual Question Answering via Temporal Token Injection and Retrieval-Augmented Adaptation\\
\noindent\textit{\textbf{Result: }} 3rd place, Source-Limited track.\\
\noindent\textit{\textbf{Members: }} \\
Yi Wen$^{1}$ (\href{mailto:guyi520421@gmail.com}{guyi520421@gmail.com}), \\
Zhenglin Du$^{1}$, \\
Zhengyang Li$^{1}$, \\
Lingling Li$^{1}$, \\
Licheng Jiao$^{1}$, \\
Wenping Ma$^{1}$\\
\noindent\textit{\textbf{Affiliations: }} \\
$^{1}$ Xidian University\\

\newpage
{    
\small     
\bibliographystyle{unsrtnat}
\bibliography{main}
}

\end{document}